\documentclass[conference]{IEEEtran}
\usepackage{graphicx}
\usepackage[hidelinks]{hyperref}
\usepackage{cite}
\usepackage{orcidlink}
\usepackage{setspace}
\usepackage{multirow}

\usepackage[skip=1ex,font=small,labelfont=bf]{caption}

\IEEEoverridecommandlockouts

\begin{document}

\title{\textbf{\LARGE \singlespace The Impact of Time Step Frequency on the Realism of Robotic Manipulation Simulation for Objects of Different Scales}}

\author{Minh Q. Ta\textsuperscript{1,\textsection}\orcidlink{0000-0002-9756-7104}, Holly Dinkel\textsuperscript{1,\textsection}\orcidlink{0000-0002-7510-2066}, Hameed Abdul-Rashid\textsuperscript{1}\orcidlink{0000-0003-2346-5989}, Yangfei Dai\textsuperscript{1}\orcidlink{0009-0006-3782-4558}, Jessica Myers\textsuperscript{1}\orcidlink{0009-0006-2824-9432}, \\ Tan Chen\orcidlink{0000-0002-4199-8706
}\textsuperscript{2}, Junyi Geng\textsuperscript{3}\orcidlink{0000-0002-6494-6810}, Timothy Bretl\textsuperscript{1}\orcidlink{0000-0001-7883-7300}
\thanks{\textsuperscript{1}University of Illinois Urbana-Champaign, Urbana, IL, 61801. \scriptsize{\texttt{{\{minh,hdinkel2,hameeda2,yangfei4,jmmyers3,tbretl\}@illinois.edu}}.}}
\thanks{\textsuperscript{2}Michigan Technological University, Houghton, MI, 49931. \scriptsize{\texttt{tanchen@mtu.edu}.}}
\thanks{\textsuperscript{3}The Pennsylvania State University, University Park, PA, 16802. \scriptsize{\texttt{jgeng@psu.edu}.}}
}

\maketitle

\begingroup\renewcommand\thefootnote{\textsection}
\footnotetext{Equal Contribution}
\endgroup

\begin{abstract}
This work evaluates the impact of time step frequency and component scale on robotic manipulation simulation accuracy. Increasing the time step frequency for small-scale objects is shown to improve simulation accuracy. This simulation, demonstrating pre-assembly part picking for two object geometries, serves as a starting point for discussing how to improve Sim2Real transfer in robotic assembly processes.
\end{abstract}

\vspace{-0.15cm}
\section{Introduction}

\begin{figure}
    \centering
    \includegraphics[width=\columnwidth]{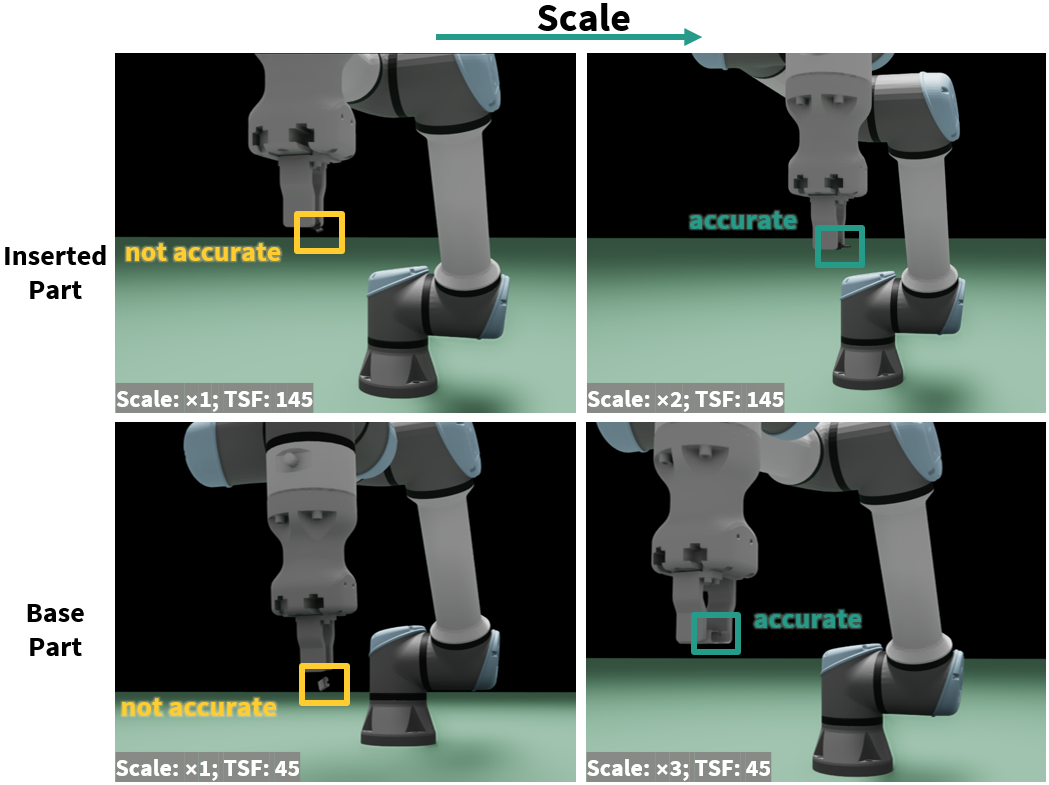}
    \caption{Inaccurate contact simulation results in differences between simulation and reality. Accurate physics constraint resolution in simulation results in reasonable component manipulation.}
    \label{fig: new_first_page}
    \vspace{-1.5em}
\end{figure}

Manipulation simulation is valuable input to the design of planning and control algorithms for real applications on robot hardware. Information about the behavior of objects in contact with each other, such as gripper fingers with a grasped component or the component with the environment, is important in the deployment of manufacturing and maintenance robots~\cite{chen2022insights,drake}. In practice, the physics of contact in hardware experiments disagrees with the physics in simulation experiments which can exhibit jittering, bouncing, sticking, penetration, or other physical constraint violations. Contact inconsistency between simulation and reality is evident when the scale of the grasped object is considered as the only variable, shown in Figure~\ref{fig: new_first_page}. Reconciling physical behaviors in simulation with behaviors in the real world (Sim2Real), is paramount for broadening adoption of robotic systems for manufacturing~\cite{heiden2020augmenting, heiden2021neuralsim}. This work evaluates the impact of Time Step Frequency (TSF) and component scale on simulation accuracy. In general, simulating any object of small volume requires increasing the TSF to achieve simulation accuracy. Simulation demonstration videos are shared publicly at \href{https://t.ly/tSlqr}{https://t.ly/tSlqr}.
\vspace{-0.2cm}
\section{Related Work}

Manipulating objects of similar shapes depends on the geometry of contact: on their sizes, on their deformability, on contact constraints with the environment, and on contact constraints with the robot. Contact and collision reduction is one of the most expensive operations performed in simulation ~\cite{erez2015sim,bouaziz2023projective,xiang2020sapien,werling2021fast,narang2022factory,mittal2023orbit}. The traditional approach to computing physics with contact constraints is to take one large time step with $n$ constraint solver iterations, solving one difficult problem accurately~\cite{baraff2023large}. More recent, real-time, local solvers, including the Temporal Gauss-Seidel (TGS) and Projected Gauss-Seidel (PGS) solvers, compute $n$ small time steps each with one constraint solver iteration, solving $n$ simpler problems approximately. Selection of a real-time local solver using smaller time steps (a higher TSF) results in fast convergence and physical feasibility of contact simulation~\cite{macklin2014unified,macklin2019small}. A manipulation policy learned from a simulation with feasible physics is more transferable to a real system.

\section{Manipulation Simulation Experiments}

\begin{figure*}
    \centering
    \includegraphics[width=\textwidth]{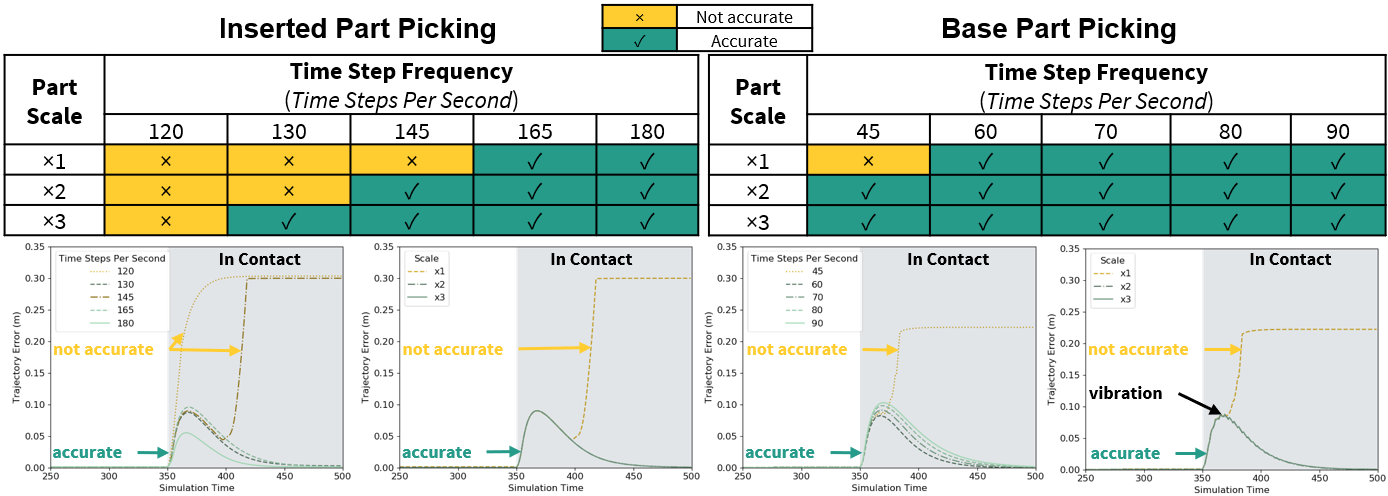}
    \caption{(Top) The simulation accuracy of picking two electronic components varies based on their scales. Objects of smaller scale require a higher Time Step Frequency (TSF) to achieve simulation accuracy. (Bottom) Trajectory error in the picking experiments was computed as the Euclidean distance between the part's center and a reference trajectory measured using the same part and scale at TSF = 360. The first and third plots represent the trajectory error for the original-scale inserted part and the base part at various TSFs. The second and fourth plots show the trajectory error for different scales of the inserted part at TSF = 145 and different scales of the base part at TSF = 45.}
    \label{fig: experiments}
    \vspace{-1em}
\end{figure*}

This work evaluates the impact of TSF, the \textit{Time Steps Per Second} simulation parameter, and component scale, measured in multiples of the original object volume and inertial properties, on manipulation simulation accuracy. The simulation environment is described in Section \ref{sec: sim-environment} and simulation results are discussed in Section \ref{sec: sim-discussion}.

\subsection{Simulation Environment}
\label{sec: sim-environment}
Manipulation simulation was performed in Isaac Sim~\cite{liang2018gpu,makoviychuk2021isaac,isaacsim}. The robot work cell in simulation included two Universal Robots UR5e manipulators, each with a Robotiq Hand-E Gripper. One manipulator was simulated performing pre-assembly part picking. The two robot movements included aligning the robot joints for grasping (non-contact) and grasping and lifting (in contact). Two components with different geometries were used in the picking task, an inserted part and a base part. Picking was performed using three different component scales, the original ($\times 1$), doubled ($\times 2$), and tripled ($\times 3$) scales. The volumes and contact surface areas of the original scales were measured in Meshlab from their associated mesh files and are shown in Table~\ref{tab: sizing}~\cite{cignoni2008meshlab}. The gripper finger contact surface area was considered to be the same as the object it grasps, so it is omitted from this table. The simulation parameters were set as \textit{Enable GPU Dynamic = True}, \textit{Solver Type = TGS}, \textit{Collider Approximation = Convex Decomposition}, \textit{Dynamic Friction = 1}, \textit{Static Friction = 1}, \textit{Restitution = 0}, and all other parameters remained default. This work used a computer with an i9 CPU, two GeForce RTX 4090 GPUs, and 64 GB RAM.

\begin{table}[h]
\footnotesize
    \centering
    \caption{Scale Comparison of Manipulation Components}
    \begin{tabular}{|c|c|c|c|}
        \hline
        Component & Inserted Part & Base Part & Gripper Finger\\
        \hline
        Volume (mm$^3$) & 178.17 & 224.29 & 13650 \\ 
        \hline
        Contact Area (mm$^2$) & 20.28 & 38.48 & - \\
        \hline
    \end{tabular}
    \label{tab: sizing}
    \vspace{-1em}
\end{table}

\subsection{Simulation Results and Discussion}
\label{sec: sim-discussion}

The experimental results for picking the two components at different scales and time step frequencies are shown in Figure~\ref{fig: experiments}. Trajectory error was measured for each scale against a reference trajectory at the same scale with 360 time steps per second. Increasing the TSF is important to improve simulation accuracy for manipulating objects of small scale. 

The trade-off between simulation accuracy and computational performance when choosing a TSF is well known. A higher TSF improves the accuracy of constraint solving, enabling more accurate resolution of contacts. However, an increased TSF increases computational cost and reduces the simulation speed. The runtime results included in Table~\ref{tab: runtime} are a reminder of this trade-off, which becomes more pronounced when simulating the manipulation of small-scale objects.

\begin{table}[h]
\footnotesize
    \centering
    \caption{Simulated Picking of Inserted Part Runtime (s)}
    \begin{tabular}{|c|c|c|c|c|c|}
        \hline
        \multicolumn{5}{|c|}{Time Steps Per Second} \\
        \hline
        120 & 130 & 145 & 165 & 180 \\
        \hline
        $10.5\pm0.2$ & $12.9\pm0.1$ & $14.8\pm0.0$ & $14.6\pm0.1$ & $18.5\pm0.1$ \\ 
        \hline
    \end{tabular}
    \label{tab: runtime}
    \vspace{-1.3em}
\end{table}

One question the results in Figure~\ref{fig: experiments} raise is whether the relationship between TSF, part scale, and simulation accuracy holds when the robot and object are close in relative size. An additional study which halved ($\times 0.5$) the robot scale for manipulation of an original-scale inserted part showed a high TSF still improves simulation accuracy, suggesting small-scale manipulation generally requires finer temporal resolution. 

\begin{figure}[h]
\vspace{-0.6em}
    \centering
    \includegraphics[width=0.75\columnwidth]{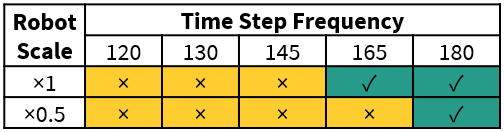}
    \caption{Increasing TSF improves manipulation simulation accuracy for inserted part picking when the robot scale is halved.}
    \label{fig: robot_scale}
    \vspace{-1.4em}
\end{figure}

\section{Conclusions and Future Work}
This work demonstrates the impact of TSF and component scale on manipulation simulation accuracy for part picking. Experiments show the selection of an appropriate number of time steps per second is essential for realistic physics, however it is one of many parameters. Future work could study the interaction of parameter combinations to further close the Sim2Real gap. The two electronic components used in this study are symmetric about one axis, however geometric symmetry may not be assumed for many types of assembly processes. For the inserted part, the center of mass is centered below its two assembly contact edges. Accounting for object geometric and inertial properties during simulation configuration can further improve Sim2Real transfer. The results from this work suggest adaptive time stepping, either through simulating different objects with different time step frequencies or adaptively sub-stepping the simulation, could balance the accuracy-performance tradeoff~\cite{soderlind2002automatic}.

\section{Acknowledgments}

\noindent \footnotesize The authors thank the members of the UIUC-FIT CoBot Factory Project and the teams developing the open-source software used in this project. M.T., H.A.-R., Y.D., J.M., T.C., J.G., and T.B. were supported by the Foxconn Interconnect Technology (FIT) and the Center for Networked Intelligent Components and Environments (C-NICE) at the University of Illinois Urbana-Champaign. H.D. was supported by the Graduate Assistance in Areas of National Need award P200A180050-19. H.D. and J.M. are supported by the NASA Space Technology Graduate Research Opportunity awards 80NSSC21K1292 and 80NSSC23K1191, respectively.

\bibliographystyle{IEEEtran}
\bibliography{IEEEabbrv,references}

\begin{thebibliography}{10}
\providecommand{\url}[1]{#1}
\csname url@samestyle\endcsname
\providecommand{\newblock}{\relax}
\providecommand{\bibinfo}[2]{#2}
\providecommand{\BIBentrySTDinterwordspacing}{\spaceskip=0pt\relax}
\providecommand{\BIBentryALTinterwordstretchfactor}{4}
\providecommand{\BIBentryALTinterwordspacing}{\spaceskip=\fontdimen2\font plus
\BIBentryALTinterwordstretchfactor\fontdimen3\font minus
  \fontdimen4\font\relax}
\providecommand{\BIBforeignlanguage}[2]{{%
\expandafter\ifx\csname l@#1\endcsname\relax
\typeout{** WARNING: IEEEtran.bst: No hyphenation pattern has been}%
\typeout{** loaded for the language `#1'. Using the pattern for}%
\typeout{** the default language instead.}%
\else
\language=\csname l@#1\endcsname
\fi
#2}}
\providecommand{\BIBdecl}{\relax}
\BIBdecl

\bibitem{chen2022insights}
T.~Chen, Z.~Huang, J.~Motes, J.~Geng, Q.~M. Ta, H.~Dinkel, H.~Abdul-Rashid,
  J.~Myers, Y.-J. Mun, W.-C. Lin, Y.-Y. Yang, S.~Liu, M.~Morales, N.~M. Amato,
  K.~Driggs-Campbell, and T.~Bretl, ``{Insights from an Industrial
  Collaborative Assembly Project: Lessons in Research and Collaboration},'' in
  \emph{{Workshop on Collaborative Robots and Work of the Future}}, 2022.

\bibitem{drake}
R.~Tedrake and the Drake Development~Team", ``{Drake: Model-Based Design and
  Verification for Robotics},'' {[Online] Available: https://drake.mit.edu},
  2019.

\bibitem{heiden2020augmenting}
E.~Heiden, D.~Millard, E.~Coumans, and G.~S. Sukhatme, ``{Augmenting
  Differentiable Simulators with Neural Networks to Close the Sim2Real Gap},''
  \emph{arXiv preprint arXiv:2007.06045}, 2020.

\bibitem{heiden2021neuralsim}
E.~Heiden, D.~Millard, E.~Coumans, Y.~Sheng, and G.~S. Sukhatme, ``{NeuralSim:
  Augmenting Differentiable Simulators with Neural Networks},'' in \emph{{IEEE}
  Int. Conf. Robot. Autom. (ICRA)}.\hskip 1em plus 0.5em minus 0.4em\relax
  IEEE, 2021, pp. 9474--9481.

\bibitem{erez2015sim}
T.~Erez, Y.~Tassa, and E.~Todorov, ``{Simulation Tools for Model-Based
  Robotics: Comparison of Bullet, Havok, MuJoCo, ODE and PhysX},'' in
  \emph{{IEEE} Int. Conf. Robot. Autom. (ICRA)}, 2015, pp. 4397--4404.

\bibitem{bouaziz2023projective}
S.~Bouaziz, S.~Martin, T.~Liu, L.~Kavan, and M.~Pauly, ``{Projective Dynamics:
  Fusing Constraint Projections for Fast Simulation},'' in \emph{Seminal
  Graphics Papers: Pushing the Boundaries}, 2023, vol.~2, pp. 787--797.

\bibitem{xiang2020sapien}
F.~Xiang, Y.~Qin, K.~Mo, Y.~Xia, H.~Zhu, F.~Liu, M.~Liu, H.~Jiang, Y.~Yuan,
  H.~Wang \emph{et~al.}, ``{Sapien: A Simulated Part-Based Interactive
  Environment},'' in \emph{{IEEE/CVF} Int. Conf. Comput. Vis. Pattern Recognit.
  (CVPR)}, 2020, pp. 11\,097--11\,107.

\bibitem{werling2021fast}
K.~Werling, D.~Omens, J.~Lee, I.~Exarchos, and C.~K. Liu, ``{Fast and
  Feature-Complete Differentiable Physics for Articulated Rigid Bodies with
  Contact},'' \emph{Robot. Sci. Syst. (RSS)}, 2021.

\bibitem{narang2022factory}
Y.~Narang, K.~Storey, I.~Akinola, M.~Macklin, P.~Reist, L.~Wawrzyniak, Y.~Guo,
  A.~Moravanszky, G.~State, M.~Lu \emph{et~al.}, ``{Factory: Fast Contact for
  Robotic Assembly},'' \emph{Robot. Sci. Syst. (RSS)}, 2022.

\bibitem{mittal2023orbit}
M.~Mittal, C.~Yu, Q.~Yu, J.~Liu, N.~Rudin, D.~Hoeller, J.~L. Yuan, R.~Singh,
  Y.~Guo, H.~Mazhar \emph{et~al.}, ``{Orbit: A Unified Simulation Framework for
  Interactive Robot Learning Environments},'' \emph{{IEEE} Robot. Autom.
  Lett.}, 2023.

\bibitem{baraff2023large}
D.~Baraff and A.~Witkin, ``{Large Steps in Cloth Simulation},'' in
  \emph{Seminal Graphics Papers: Pushing the Boundaries}, 2023, vol.~2, pp.
  767--778.

\bibitem{macklin2014unified}
M.~Macklin, M.~M{\"u}ller, N.~Chentanez, and T.-Y. Kim, ``{Unified Particle
  Physics for Real-Time Applications},'' \emph{{ACM} Trans. Graph. (TOG)},
  vol.~33, no.~4, pp. 1--12, 2014.

\bibitem{macklin2019small}
M.~Macklin, K.~Storey, M.~Lu, P.~Terdiman, N.~Chentanez, S.~Jeschke, and
  M.~M{\"u}ller, ``{Small Steps in Physics Simulation},'' in \emph{{ACM
  SIGGRAPH/Eurographics Symp. Comput. Animat.}}, 2019, pp. 1--7.

\bibitem{liang2018gpu}
J.~Liang, V.~Makoviychuk, A.~Handa, N.~Chentanez, M.~Macklin, and D.~Fox,
  ``{GPU-Accelerated Robotic Simulation for Distributed Reinforcement
  Learning},'' in \emph{Int. Conf. Robot Learn. (CoRL)}, 2018, pp. 270--282.

\bibitem{makoviychuk2021isaac}
V.~Makoviychuk, L.~Wawrzyniak, Y.~Guo, M.~Lu, K.~Storey, M.~Macklin,
  D.~Hoeller, N.~Rudin, A.~Allshire, A.~Handa \emph{et~al.}, ``{Isaac Gym: High
  Performance GPU-Based Physics Simulation for Robot Learning},'' \emph{arXiv
  preprint arXiv:2108.10470}, 2021.

\bibitem{isaacsim}
NVidia, ``{Isaac Sim - Robotics Simulation and Synthetic Data Generation},''
  [Online] Available: https://developer.nvidia.com/isaac-sim, 2023.

\bibitem{cignoni2008meshlab}
P.~Cignoni, M.~Callieri, M.~Corsini, M.~Dellepiane, F.~Ganovelli, G.~Ranzuglia
  \emph{et~al.}, ``{Meshlab: An Open-Source Mesh Processing Tool},'' in
  \emph{Eurographics Italian Chapter Conference}, vol. 2008.\hskip 1em plus
  0.5em minus 0.4em\relax Salerno, Italy, 2008, pp. 129--136.

\bibitem{soderlind2002automatic}
G.~S{\"o}derlind, ``{Automatic Control and Adaptive Time-Stepping},''
  \emph{Numer. Algorithms}, vol.~31, pp. 281--310, 2002.

\end{thebibliography}

\end{document}